\documentclass[journal]{IEEEtran}

\usepackage{graphicx}
\usepackage{subfigure}
\usepackage[numbers,sort&compress]{natbib}

\usepackage{caption2}
\usepackage{amsmath}
\usepackage{array,diagbox,siunitx}

\usepackage{cases}
\usepackage{float}
\usepackage{amssymb}
\usepackage{multirow}
\usepackage{mathrsfs}
\usepackage{algorithmic}
\usepackage[ruled]{algorithm2e}
\usepackage{rotating}
\usepackage{balance}
\usepackage{threeparttable}
\usepackage{color}
\usepackage{colortbl}
\definecolor{mygray}{gray}{.9}
\usepackage{makecell}
\usepackage{pdflscape}
\usepackage{siunitx}
\usepackage{stfloats}
\usepackage{verbatim}
\usepackage{setspace}
\usepackage{threeparttable}
\usepackage{supertabular,booktabs}
\usepackage{rotating}
\usepackage{supertabular}
\usepackage{pifont}
\usepackage{xcolor,pict2e}

\usepackage{tikz}
\usepackage{booktabs}
\usepackage{footnote}
\usepackage[pagewise]{lineno}
\usepackage{balance}
\newcommand{\etal}{\textit{et al.}}
\newcommand{\ie}{\textit{i.e.}}
\newcommand{\eg}{\textit{e.g.}}

\begin{document}
\title{Region attention and graph embedding network for occlusion objective class-based micro-expression recognition}
\author{Qirong Mao,~\IEEEmembership{Member,~IEEE,}
        Ling Zhou,~\IEEEmembership{Student Member,~IEEE,}
        Wenming Zheng,~\IEEEmembership{Senior Member,~IEEE,}
        Xiuyan Shao,
        Xiaohua Huang,~\IEEEmembership{Member,~IEEE}
\thanks{Q. Mao and L. Zhou are with the School of Computer Science and Communication Engineering, Jiangsu University, Zhenjiang, Jiangsu, China.
(e-mail: mao\_qr@ujs.edu.cn, 2111808003@stmail.ujs.edu.cn)}
\thanks{W. Zheng is with the Key Laboratory of Child Development and Learning Science (Southeast University), Ministry of Education, Southeast University, Nanjing 210096, China and also with the School of Biological Science and Medical Engineering, Southeast University, Nanjing 210096, Jiangsu, China.
(E-mail: wenming\_zheng@seu.edu.cn.)}
\thanks{X. Shao is with Southeast University, Nanjing 210096, Jiangsu, China.
(E-mail: xiuyan\_shao@seu.edu.cn.)}
\thanks{X. Huang is with School of Computer Engineering, Nanjing Institute of Technology, China.
(email: xiaohuahwang@gmail.com.)}
}

\markboth{2021}%
{Shell \MakeLowercase{\textit{et al.}}: Bare Demo of IEEEtran.cls for IEEE Journals}

\maketitle
\begin{abstract}
Micro-expression recognition (\textbf{MER}) has attracted lots of researchers' attention in a decade. However, occlusion will occur for MER in real-world scenarios. This paper deeply investigates
an interesting but unexplored challenging issue in MER, \ie, occlusion MER. First, to research MER under real-world occlusion, synthetic occluded micro-expression databases are created by using various mask for the community. Second, to suppress the influence of occlusion, a \underline{R}egion-inspired \underline{R}elation \underline{R}easoning \underline{N}etwork (\textbf{RRRN}) is proposed to model relations between various facial regions. RRRN consists of a backbone network, the Region-Inspired (\textbf{RI}) module and Relation Reasoning (\textbf{RR}) module. More specifically, the backbone network aims at extracting feature representations from different facial regions, RI module computing an adaptive weight from the region itself based on attention mechanism with respect to the unobstructedness and importance for suppressing the influence of occlusion, and RR module exploiting the progressive interactions among these regions by performing graph convolutions. Experiments are conducted on handout-database evaluation and composite database evaluation tasks of MEGC 2018 protocol. Experimental results show that RRRN can significantly explore the importance of facial regions and capture the cooperative complementary relationship of facial regions for MER. The results also demonstrate RRRN outperforms the state-of-the-art approaches, especially on occlusion, and RRRN acts more robust to occlusion.

\end{abstract}


\IEEEpeerreviewmaketitle

\section{Introduction}
\IEEEPARstart{M}{icro-expressions}, different from macro-expressions or subtle expressions, \textcolor{black}{are hidden emotions with low intensity and frag-mental facial action units where only part of the facial muscles of full-stretched facial expressions are presented \cite{Porter2008}}. Micro-expression recognition (\textbf{MER}) aims at revealing the hidden emotions of humans and understanding people's deceitful behaviors when micro-expressions occur \cite{Michael2010, Ekm2009}.
As it holds tremendous potential impact on a wide-range of applications including psychology, medicine, and police case diagnosis \cite{Ekman1969, Ekm2009}, MER has attracted significant interest from psychologists and computer scientists in recent years.


Although many MER approaches have been proposed, all of them are built on samples captured in laboratory-controlled environments, such as \textcolor{black}{CASME II} \cite{Yan2014} and SAMM \cite{Davison2018}. Unfortunately, in many real-world scenarios, the human's face may be partially occluded by such as glasses. For example, a sunglass or a virtual reality mask occludes eyes, while a scarf or medical mask occludes the mouth and nose. especially in the challenging Coronavirus epidemic prevention period. Thus far, occlusion becomes a challenging problem in the field of face analysis task since it
causes noise to feature descriptor. \textcolor{black}{Note that,} occlusion problem has been received attention in face identity recognition \cite{Yang2017,Xu2020} and macro-expression recognition \cite{Wang2020b,Li2019d}. However, little attention has been paid to occlusion for MER, due to the lack of occluded micro-expression databases and efficient occlusion MER methods. \textcolor{black}{In last decade,} most of current MER methods have focused on micro-expression without occlusion. Our empirical experience \textcolor{black}{in Section~\ref{sec:experimentD}} shows these methods perform unsatisfactorily for MER on partial occlusion. Therefore, it is very worthwhile to investigate the influence of occlusion to MER and propose a more robust method for MER to resolve occlusion problem.


To achieve the above-mentioned goals, we first construct six occlusion micro-expression databases derived from CASME II and SAMM by considering different occlusion cases, namely Mask-CASME II, Glass-CASME II, Random Mask CASME II, Mask-SAMM, Glass-SAMM, \textcolor{black}{and} Random Mask SAMM. The occlusion databases are manually synthesized with occlusion types of wearing masks, glasses, and random region masks on micro-expression sequences.
Second, motivated by~\cite{Csurka2004,Xie2019a, Wang2020c} in image understanding, we propose the Region-inspired Relation Reasoning Network (\textbf{RRRN}) for MER to highlight the important facial regions and model region relationships for enhancing the ability of resistance to occlusion. It consists Backbone network, Relation Reasoning (\textbf{RR}) module, and Region-Inspired (\textbf{RI}) module. Briefly, Backbone network obtains the coarse region features, RR module learns the global feature by building relationship on the complementary region feature, RI module aims at obtaining the weighted region features for RR module. Moreover, with specific data augmentation strategy for micro-expression samples, experimental results demonstrate that our proposed RRRN substantially \textcolor{black}{suppresses} the influence of occlusion to MER and outperforms the existing state-of-the-art approaches in both partially occluded and non-occluded situations.


The four contributions of our work are summarized as follows.
\begin{itemize}
\item We synthesize a number of occluded micro-expression databases. To the best of our knowledge, it is the first to build micro-expression databases in occlusions for video-based MER.

\item We present a Region-inspired relation reasoning network to resolve occlusion problem of MER by incorporating region weighted feature learning and region-based relation reasoning into embedding learning in a deep learning network. 

\item We propose a Region-Inspired module to compute an adaptive weight from the region itself according to the unobstructedness and importance of region. We also introduce a novel region graph representation to capture relationships between attended parts in a single micro-expression sample. Moreover, graph convolutional network-based parts relation reasoning on this graph is then performed, \textcolor{black}{leading} to the complementary Relation Reasoning module.

\item Experimental results demonstrate the proposed RRRN yields other state-of-the-art methods \textcolor{black}{on non-occluded and synthesized partially occluded micro-expression databases} under the MEGC 2018 settings.
\end{itemize}

\section{Related Work}
Since the occlusion MER problem has not yet been investigated, in this section we literature previous works closely related to ours,~\ie, features in MER and \textcolor{black}{occluded} facial expression recognition.

\subsection {Features in MER}




Most handcrafted features based methods are originally designed for MER. According to \cite{Oh2018a}, they are divided into appearance-based and geometric-based features. The widely used representation is Local Binary Pattern from Three Orthogonal Planes (LBP-TOP) \cite{Zhao2007}. Considering its low computational complexity,  many LBP-TOP variants have been proposed, \textcolor{black}{\eg,} Spatiotemporal Completed Local Quantized Patterns (STCLQP) \cite{Huang2016}, hierarchical spatiotemporal descriptors \cite{Zong2018a}, discriminative spatiotemporal LBP with revisited integral projection (DiSTLBP-RIP) \cite{Huang2019}  and others \cite{Wang2015, Wang2014, Wang2016, Wang2015a, Huang2015}. Besides the LBP family, 3D Histograms of Oriented Gradients (3DHOG) \cite{Polikovsky2009, Polikovsky2013} focuses on counting occurrences of gradient orientation in localized portions of the image sequence.
Different from appearance-based features, geometric-based features aim to represent micro-expression samples by the aspect of face geometry, \textcolor{black}{\eg,} shapes and location of facial landmarks. The representations include Delaunay-based Temporal Coding Model (DTCM) \cite{Lu2014}, Main Directional Mean Optical Flow (MDMO) \cite{Liu2016}, Facial Dynamics Map (FDM) ~\cite{Xu2017}, and Bi-Weighted Oriented Optical Flow (Bi-WOOF) ~\cite{Liong2018a}.

Besides the handcrafted features, many feature learning based methods have been proposed for MER \cite{Kim2016,Li2018, Reddy2019,Xia2020,Verma2020a,Wang2020,Verma2020, Li2021}.
For example, to better represent the subtle changes in micro-expression, \textcolor{black}{Li~\etal~\cite{Li2021} proposed a joint feature learning architecture
coupling local and global information for MER.}
Khor~\etal~\cite{Khor2018} adopted Enriched Long-term Recurrent Convolutional Network based on optical flow features, which \textcolor{black}{containes} the channel-wise spatial enrichment and the feature-wise temporal enrichment. Meanwhile, Li \etal~\cite{Li2018} claimed three-stream 3D flow convolutional neural network and Peng~\etal~\cite{Peng2017} leveraged Dual Temporal Scale Convolutional Neural Network (DTSCNN) for MER.
The DTSCNN was the first work in MER that utilized a shallow two-stream neural network with inputs of optical-flow sequences. Then, several shallow networks were proposed for MER~\cite{Gan2019,Liong2019,Zhou2019,Liu2019,Quang2019, Liong2019a, Xia2020a, Xie2020, Lei2020} and they avoided from the over-fitting problem caused by the scarcity of micro-expression data. Different from those models which used shallow networks to alleviate over-fitting, Peng~\etal~\cite{Peng2018} adopted pre-trained ResNet10 \cite{Simon2016} as a backbone and introduced  transfer learning strategy on macro-expression databases to handle the over-fitting issue.
Recently, motivated by the attention strategy and the transfer learning mechanism, Zhou \etal~\cite{Zhou2019a} utilized ResNet with the input of apex frames for MER. More recently, to explore the semantic relationships between Action Units and emotion classes, Lo \etal~\cite{Lo2020} and Xie \etal~\cite{Xie2020} used graph convolutional network (GCN) \cite{Tang2020} to MER. Compared with other deep methods, both works~\cite{Wang2020,Xia2020} indicate that Eulerian Video Magnification (EVM) contributes to increasing recognition accuracy. However, it is hard to control the magnification factor of EVM in real-world scenarios. Xia~\etal ~\cite{Xia2020} proposed an effective data augmentation method to avoid from the over-fitting problem and a balanced loss function to tackle the data imbalance issue for MER. Additionally, Yu~\etal~\cite{Yu2020} proposed a novel Identity-aware and Capsule-Enhanced Generative Adversarial Network to improve MER performance in an end-to-end way.

\textcolor{black}{From the above features learning in MER, it is clear they all focused on un-occluded MER, and ignoring digging the different importance of facial region features for MER,  let alone occlusion MER.  Our work differs from these approaches in that it explicitly learns weighted facial region features and capture
the cooperative complementary relationship of facial regions for occlusion MER.}

\subsection{Occluded facial expression recognition}

Facial expression recognition (FER) research on handling occlusion has been studied over the past decade. \textcolor{black}{Amounts} of deep learning methods have been proposed to resolve partial occlusions for FER \cite{Wang2020b,Hu2020,Houshm2020,Ding2020,Georgescu2020, Li2019d,Dong2019}. For example, Houshm~\etal~leveraged a transfer learning methods to get more discriminative features in presence of occluded facial region, when the user is wearing a head-mounted display in a VR setting~\cite{Houshm2020}.
Hu \etal~tackled the partial occlusion for FER by using face inpainting and region features~\cite{Hu2020}. More specifically, the symmetric SURF and face inpainting with mirror transition are applied to detect occluded part, such that it reduced the influence of occluded areas on the performance. Subsequently, a FER network based on heterogeneous soft blocks are leveraged to weight the importance of each area. Li \etal~presented a convolutional neutral network with attention mechanism to concern about the contribution of different facial regions~\cite{Li2019d}. More recently, Ding \etal~proposed Occlusion-Adaptive Deep Network (OADN) with a landmark-guided attention branch and a facial region branch~\cite{Ding2020}. In \textcolor{black}{that work}, landmark-guided attention branch \textcolor{black}{discards} feature elements which have been corrupted by occlusions, while the facial region branch \textcolor{black}{learns} robust feature with complementary context information. Wang \etal~proposed Region Attention Network (RAN) to learn robust representation for occlusion FER by exploring combination of local and global features~\cite{Wang2020b}. RAN adaptively captures the importance of facial regions for occlusion and pose variant FER by leveraging the self-attention mechanism, and aggregating those weighted region features into a global representation for final prediction.

It is worth mentioning that RAN is the most related work to ours, but our work is distinct from RAN~\cite{Wang2020b}. The differences are presented as follows: (1) We propose a new method to learn region features. Wang~\etal~\cite{Wang2020b} used a simple attention mechanism to obtain the region feature and did not consider the information decay. \textcolor{black}{By contrast}, we propose a Region-Inspired (RI) module with region attention to better obtain the region information. (2) We propose the different strategy to get global representation. In~\cite{Wang2020b}, global feature is concatenated by the refined region features with relation-attention module but ignoring the complementary relationship of regions. Different from~\cite{Wang2020b}, we leverage RR module to capture appearance relationships among the weighted facial regions by GCN reasoning. The outputs of such GCNs are \textcolor{black}{the} updated node features (with each node representing an attended facial region), which are further used to learn embedding to the semantic space and to recover missed features for occlusion MER; (3) Besides only the region biased loss used in \cite{Wang2020b} to capture the importance of regions, a correlation loss in our work is introduced to enhance the robustness of global feature by balancing the relationship between local and global representation.

\section{METHODOLOGY}
\label{sec:method}

In this section, we first overview the proposed RRRN, and then describe its architecture in detail. Finally, we elaborate on the usage of loss functions for RRRN.

\subsection{Overview}


RRRN architecture mainly consists of three important modules: the Backbone network, the Region-Inspired module (RI), and the Relation Reasoning (RR) module. To suppress the influence of occlusion to MER, RRRN endeavours to adaptively extract the importance of facial regions in the RI module, and reasonably to model complementary relationships between different facial regions \textcolor{black}{for learning} robust features in the RR module. All the modules are jointly trained by the objective function loss (Eqn.~\ref{loss_all}), which is composed of a region biased loss (Eqn.~\ref{loss_rb})~\cite{Wang2020b}, a cross-entropy loss (Eqn.~\ref{loss_cls}), and an introduced correlation loss (Eqn.~\ref{loss_cor})~\cite{Wang2020c}.  In the pre-processing, for each micro-expression sequence, we calculated the TV-L1 \cite{Zach2007} optical flow \textcolor{black}{from} the onset and apex frames. Motivated by \cite{Wang2020b}, given horizontal and vertical components of each optical flow, we first \textcolor{black}{crop} them into a number of regions with fixed position cropping. Subsequently, the cropped regions along with the original optical flow region are fed into the Backbone where its outputs are some region features. Next, each region is assigned with an attention weight by using RI. Afterward, RR represents the region features into relational region features, by reasoning relationship among individual region features to further capture content-aware global graph embedding. Lastly, we \textcolor{black}{leverage} the weighted region featured and the global graph representation to predict the micro-expressions.

\subsection{Motion information feature}
As a motion information feature, the optical flow is extensively used by~\cite{Liong2019,  Gan2019, Liong2019b} for micro-expression recognition. More specifically, the optical flow of each sample is extracted from the onset and apex frames, where onset and apex frames mean the frames with neutral-expression and the highest expression intensity, respectively. For RRRN, TV-L1 optical flow method~\cite{Zach2007} is utilized to obtain motion features from the onset and apex frames of each micro-expression video.

For preserving more motion information, two optical flow components are extracted to represent the facial change along horizontal and vertical directions. Then, to capture different region features, we \textcolor{black}{use} the fixed position cropping method \cite{Wang2020b} on optical flows to generate facial optical flow regions. Specifically, five regions are cropped, where three of them are the top-left, top-right, and center-down face regions, which have a fixed size of 0.75 scale ratio of the original face, and the other two regions are the center regions with sizes of 0.9 and 0.85 scale ratio of the original face. These five crops facial optical flow regions along with their corresponding original facial region are resized into the same size for the input of the Backbone network. Each region is a composite of a vertical component and a horizontal component of the optical flow. Formally, we denote each sample as $x$, and the vertical and horizontal components of the duplicated original region as $x_v^0$ and $x_h^0$, respectively.
Subsequently, the input of the whole model is defined by:
\begin{equation}
\boldsymbol{X} = \{[x_v^0,...,x_v^k,..., x_v^K], [x_h^0,...,x_h^k,..., x_h^K]\},
\label{input}
\end{equation}
where $k$ represents the $k$-th facial region, and $0 \leq k \leq K$, $K=5$.

\subsection{Backbone network}

$K$ facial regions are fed into a CNN module with two ResNet18 and a concatenated layer. In detail, given the input $\boldsymbol{X}$, the vertical and horizontal feature maps are extracted by the CNN module, where they are denoted as $[Z_1(x_v^0), ..., Z_1(x_v^K)]$ and $[Z_2(x_h^0),..., Z_2(x_h^K)]$, respectively.
Then, the two components feature maps corresponding to the $k$-th facial region are concatenated into one feature map $p_k = [Z_1(x_v^k); Z_2(x_h^k))]$ to represent the $k$-th region.

\subsection{Relation Reasoning (RR) module}


According to~\cite{Ekman1978}, FACS demonstrates that there exists relationship between facial muscles. To imitate the relationship between facial regions, we develop RR module based on GCN to model the relationship between facial regions. Experiments of model ablation in Section~\label{sec:experimentC} validates that our developed RR module helps RRRN boost the performance.



In detail, we first construct a region graph $\Gamma \in \mathbb{R}^{K\times K}$ with $K$ region features as its $K$ nodes. The dot-product is leveraged to calculate the pairwise similarity:
\begin{equation}
\Gamma_{ik} = <p_i, p_k>, 0\leq i \leq K, 0\leq k \leq K.
\label{mu}
\end{equation}
where $f_k$ is $L_2$-normalized, and $p_i$ and $p_k$ mean the $i$-th and $k$-th region features, respectively.

As the dot-product calculation is equivalent to the cosine similarity metric and the graph has self-connections, the degree matrix $\mathbf{D}$ of $\Gamma$ is calculated as followed:
\begin{equation}
D_{kk} = \Sigma_{i=1}^K \Gamma_{ki}.
\label{mu2}
\end{equation}

Given $\Gamma$, we \textcolor{black}{leverage} GCN method to perform reasoning on this graph. In GCN, graph convolution allows each target node in the graph to aggregate features from all neighbor nodes according to the edge weight between them. It means that messages can be passed inside the graph to update each AU feature. Therefore, the GCN outputs can \textcolor{black}{be used to} update relational features of each region node. Formally, one GCN layer is represented as,
\begin{equation}
\footnotesize
\begin{split}
\mathbf{P}^{(l+1)} = \sigma (\mathbf{D}^{-1}\Gamma \mathbf{P}^{(l)}\mathbf{W}^{(l)}), l=0,1\\
\end{split}
\label{gcn}
\end{equation}
where $\sigma(.)$ denotes the ReLU activation function in our work, and $\mathbf{P}^{(0)} \in \mathbb{R}^{K\times C}$ denotes the stacked $K$ region features, $C$ is their dimensions. $\mathbf{W}^{(l)}$ is the learnable weight matrix. Subsequently, the $K$ region features in  $\mathbf{P}^{(2)} \in \mathbb{R}^{K\times C}$ are updated by GCNs. Lastly, the $K$ region features are aggregated for classification.

\subsection{Region-Inspired (RI) module for suppressing occlusion}

Although our previously developed RR module efficiently learns the relationship between facial regions, RR has weakness to address the occlusion situation when occlusion coming. Therefore, it is expected that one additional module can better polish the noise caused by occlusion for RR module.

Considering the above-mentioned expectation, we propose Region-Inspired (RI) module, which aims at automatically learning low weights for the occluded regions but \textcolor{black}{also} high weights for the un-occluded and discriminative regions. It consists of region-squeeze and concatenating, region attention, and weighted region feature stages. The first stage encodes feature maps of each input region by using an independent region-squeeze block to preserve more information when learning region specific patterns. The second stage adaptively learns the weights of different facial regions according to their contribution to the recognition. The last stage obtains the weighted region feature maps by multiplying the region features with its corresponding learned weights.


Region feature maps are first fed into a convolution layer without decreasing the spatial resolution, then  pooling beside the channel axis are applied, following with another convolution layer.
We \textcolor{black}{use} max-pooling and average-pooling besides channel axis and \textcolor{black}{concatenate} both to obtain feature descriptors. Mathematically speaking, suppose that $p_k \in \mathbb{R}^{H\times W \times C}$ denotes the input feature maps of the $k$-th region, the $k$-th region-squeeze block takes the feature maps $p_k$ as the input and learns the squeezed region feature maps $\mu_k \in \mathbb{R}^{H\times W \times 1}$:
\begin{equation}
\mu_k = F_2([AvgPool(F_1(p_k)); MaxPool(F_1(p_k))]),
\label{ms}
\end{equation}
 where $F_1$ and $F_2$ mean the convolution layer before and after pooling, respectively. $[\cdot]$ is feature concatenating operation.

Afterwards, $K$ sets of squeezed region feature maps are concatenated into one set of feature maps $\psi \in \mathbb{R}^{H\times W \times K}$, with $H$, $W$, and $K$ being its height, width, and the number of regions, respectively.

\subsection{Embed RI into RR module and feature aggregation}

With the weighted feature $f_k$, we can re-formulate Eqns.~(\ref{mu}) and~(\ref{gcn}) by replacing $p_k$ with $f_k$. Lastly, the $K$ weighted region features in $\mathbf{F}^{(2)} \in \mathbb{R}^{K\times C}$ updated by GCNs and the $K$ weighted region features $f_k (0\leq k\leq K$) further undergo an element-wise addition, respectively. Therefore, RI is embeded into RR module, such that it assist RR module to suppress the influence of occlusion for MER.

Before aggregating features for classification, weighted region features and relational region features are concatenated to generate micro-expression-level representations for final classification.

\subsection{Loss design of RRRN}
The objective of RRRN is to correctly classify the micro-expression samples into their corresponding labels $y$. The standard cross-entropy loss is employed for micro-expression predication,
\begin{equation}
\mathcal{L}_{cls}=-\sum_{i=1}^{N}[y_i\log(\frac{exp(\Phi_i)}{\sum_i {exp(\Phi_i)}})],  \\
\label{loss_cls}
\end{equation}
where $N$ is the number of the training samples, $y_i$ is the objective class label of the $i$-th training instance, $\Phi_i$ denotes the last fully connected layer in the model activated by a $softmax$ unit.

 As aforementioned, RI is a local feature learning stage while RR is a global feature learning stage. More specifically, RI aims to learn region features with different weights according to the contribution for perceiving the occluded facial regions, and RR aims at learning more effective representation by capturing the cooperative complementary relationship of facial regions. To learn better region features and more robust global feature, two additional losses are thus introduced in both two modules for considerable performance.

\subsubsection{\textbf{Region Biased Loss}}

\textcolor{black}{According to \cite{Boucher1975}, different facial expressions are mainly defined by facial action units. It is encouraged to assign a high attention weights to the most import region. Therefore, Region Biased Loss (RB-Loss)~\cite{Wang2020b} is used in our RI module to make a straightforward constraint on the attention weights. It resorts a simple constraint in \cite{Wang2020b} that the maximum attention weight of facial regions should be larger than the one of the original face image. In other words, RB-Loss can make an agreement with our goal. The RB-Loss is formulated as:}
\begin{equation}
\begin{split}
\mathcal{L}_{rb} = max\{0, \beta -(\alpha_{max}-\alpha_0)\},
\end{split}
\label{loss_rb}
\end{equation}
where $\beta$ is a hyper-parameter, $\alpha_{max}$ is the maximum attention weight of all the $K-1$ facial crops, and $\alpha_0$ denotes the attention weight of the original face region. RB-Loss restrains the relationship among region features by using region feature learning based on prior knowledge.

\subsubsection{\textbf{Correlation Loss}}

\textcolor{black}{The Bag of Words \cite{Csurka2004} indicates the aggregated feature obtains the higher prediction probability than single region feature. Motivated by the work \cite{Csurka2004},} we embed the Correlation loss (Cor-Loss) into RI and RR modules in RRRN  for restraining the relationship between region features and global features. \textcolor{black}{Importantly,} the Cor-Loss can guarantee that the prediction probability of the final combined feature is greater than that of single region features. \textcolor{black}{It} is formulated as followed:
\begin{equation}
\begin{split}
\mathcal{L}_{cor} = \sum (max\{0, log P(f_k)-log P(f_a)\}),
\end{split}
\label{loss_cor}
\end{equation}
where $f_k$ denotes the \textcolor{black}{feature of} the $k$-th facial region, the function $P$ is the confidence function which reflects the probability of classification into the correct category, $f_a$ is the aggregation of weighted region features and relational region features.

\subsubsection{\textbf{Final loss function}}

The final loss function is defined as followed:
\begin{equation}
\mathcal{L}= \mathcal{L}_{cls} + \lambda_1\mathcal{L}_{rb}+ \lambda_2\mathcal{L}_{cor},
\label{loss_all}
\end{equation}
where $\lambda_1$ and $\lambda_2$ balance these losses.

\section{EXPERIMENTS}
\label{sec:experiment}
\textcolor{black}{We focus on objective class-based MER based on the MEGC 2018 protocol} for our experiment. In this section, we will first describe the experiment settings, including the two tasks of objective class-based MER, evaluation metrics, and our implementation details. Then, we compare our method with the state-of-the-art objective class-based MER methods on un-occluded and synthesized occluded databases. Finally, feature maps and attention weights of the facial regions in the RI module are visualized for analysis.

\subsection{Un-occluded and occluded micro-expression Databases}

\subsubsection{Un-occluded micro-expression databases}
In this paper, we focus on MEGC 2018 setting~\cite{Yap2018} for micro-expression recognition  on objective classes. \textcolor{black}{Specifically, objective classes are defined in \cite{Davison2018a} according to Facial Action Coding System (FACS) \cite{Ekman1978}. The relationship between action units and objective classes I-V is shown in Table \ref{objective_class}.} MEGC 2018 setting \textcolor{black}{combines} CASME II \cite{Yan2014} and SAMM \cite{Davison2018} datasets into a composite database. It primarily focuses on the first five classes,~\ie,~objective classes I to V.  The composite database contains 185 samples of 26 subjects in CASME II and 68 samples of 21 subjects in SAMM, respectively. Table \ref{data} summarizes the composite database and its sample distribution over each objective class.

\begin{table}
\caption{Relationship between action unit and objective classes I-V.}
\label{objective_class}
\begin{center}
\setlength{\tabcolsep}{0.9mm}{
\begin{tabular}{|c|l|}
\hline
Class & Action Units \\
\hline
I&AU6, AU12, AU6+AU12, AU6+AU7+AU12, AU7+AU12\\
\hline
\multirow{2}*{II}&AU1+AU2, AU5, AU25, AU1+AU2+AU25, AU25+AU26,\\
&AU5+AU24\\
\hline
\multirow{2}*{III}&A23, AU4, AU4+AU7, AU4+AU5, AU4+AU5+AU7,\\
&AU17+AU24, AU4+AU6+AU7, AU4+AU38\\
\hline
\multirow{3}*{IV}&AU10, AU9, AU4+AU9, AU4+AU40, AU4+AU5+AU40,\\
&AU4+AU7+AU9, AU4 +AU9+AU17, AU4+AU7+AU10,\\
&AU4+AU5+AU7+AU9, AU7+AU10\\
\hline
V&AU1, AU15, AU1+AU4, AU6+AU15, AU15+AU17\\
\hline
\end{tabular}}
\end{center}
\end{table}

\begin{table}[h]
\caption{Sample information of objective classes I-V in CASME II and SAMM datasets.}
\label{data}
\begin{center}
\setlength{\tabcolsep}{1.5mm}{
\begin{tabular}{|c|c|c|c|c|c|c|c|}
\hline
\multirow{2}*{Database}&\multicolumn{5}{c|}{Objective Class}&\multirow{2}*{Total}&\multirow{2}*{Subjects}\\
\cline{2-6}
&I&II&III&IV&V&&\\
\hline
\hline
CASME II \cite{Yan2014}&25&15&99&26&20&185&26\\
\hline
SAMM \cite{Davison2018}&24&13&20&8&3&68&21 \\
\hline
Composite&49&28&119&34&23&253&47\\
\hline
\end{tabular}}
\end{center}
\end{table}

\subsubsection{Synthesis of occluded micro-expression databases}

As there is no occluded micro-expression database available so far, we synthesized various occlusion cases on CASME II and SAMM databases for our experiments. More specifically, we synthesized with occlusion types of wearing masks, glasses, and random region masks (5\%-50\% partial occlusions) on un-occluded micro-expression sequences. Therefore, six occlusion micro-expression databases were derived, namely Mask-CASME II, Glass-CASME II, Random Mask CASME II (RMask-CASME II), Mask-SAMM, Glass-SAMM, and Random Mask SAMM (RMask-SAMM).

In daily life, the head movements and facial changes usually lead to the movements of some accessories, especially masks and glasses. For example, the tilted head causes the masks leaned at the same angle, and some expression related to the movement of frowning may cause a slight shift in the position of the sunglasses. In order to make the synchronous movement of accessories and face, the occlusion patch is fixed in the specific region according to facial landmarks in each frame.
Specifically, We manually collected ten masks and ten sunglasses from the website as the occlusion patches. Subsequently, we generated the location of masks, sunglasses, and random region masks for each frame in a micro-expression sequence according to the detected landmarks.

\subsection{Experimental settings}
\textbf{(1) Task of MEGC 2018}

 MEGC 2018 contains two \textcolor{black}{tasks}: Holdout-database evaluation (HDE) and Composite database evaluation (CDE) tasks. In HDE task, there is 2-fold cross-validation in HDE protocol, ~\ie, training on CASME II while testing on SAMM (\emph{CASME II$\rightarrow$SAMM}), vice versa (\emph{SAMM$\rightarrow$CASME II}). Unweighted average recall (UAR) and weighted average recall (WAR) \cite{Schuller2010} are introduced in the task to measure the performance of different approaches. The average WAR and UAR are used in the experiment. In CDE task, Leave-One-Subject-Out (LOSO) cross-validation protocol is used. All samples from CASME II and SAMM \textcolor{black}{databases} are combined into a single composite database. Samples of each subject are held out as the testing set while \textcolor{black}{the rest} for training. F1 score and Weighted F1 score (WF1) are used to measure the performance of various methods. Here, the F1 is an average of the class-specific F1 across the whole classes (or macro-averaging), and WF1 is weighted by the number of samples in \textcolor{black}{their} corresponding classes before averaging \cite{Yap2018}. These metrics are calculated as follows:
\begin{equation}
\footnotesize
WAR = \frac{\sum_{c=1}^{C}TP_c}{N},
\label{war}
\end{equation}
\begin{equation}
\footnotesize
UAR = \frac{1}{C}\sum_{c=1}^{C}\frac{TP_c}{N_c},
\label{uar}
\end{equation}
\begin{equation}
\footnotesize
F1 = \frac{1}{C}\sum_{c=1}^{C}\frac{2\cdot TP_c}{2\cdot TP_c + FP_c+FN_c},
\label{f1}
\end{equation}
\begin{equation}
\footnotesize
WF1 = \sum_{c=1}^{C}\frac{N_c}{N}\frac{2\cdot TP_c}{2\cdot TP_c + FP_c+FN_c},
\label{f1}
\end{equation}
where $C$ is the number of classes, $c\leq C$, $N_c$ is the number of samples with the $c-$th class, and  $N$ is the total samples.
$TP_c$,  $FP_c$,  and $FN_c$ are the  true positives,  false positives, and  false negatives of the $c-$th class, respectively.

\textbf{(2) Implementation details}

In the pre-processing step, we \textcolor{black}{utilize} face\_recognition algorithm~\footnote{https://github.com/ageitgey/face\_recognition} to obtain the facial area of each frame, and then \textcolor{black}{extracte} TV-L1 optical flow \cite{Zach2007} features from the onset and apex frames. Moreover, two components of optical flow images \textcolor{black}{are} resized to $224 \times 224$ \emph{pixels}.



\textbf{Data augmentation:} Due to limited sample in micro-expression database, data augmentation \textcolor{black}{is} applied to our proposed network. For each micro-expression video sequence, the position of onset, apex and offset frames \textcolor{black}{are} denoted as ${\textit{Pos}}_{\textit{onset}}$, ${\textit{Pos}}_{\textit{apex}}$ and ${\textit{Pos}}_{\textit{offset}}$, respectively. First, 4 frames in the position of ${\textit{Pos}}_{\textit{onset}} + ({\textit{Pos}}_{\textit{apex}} - {\textit{Pos}}_{\textit{onset}}) \times [0.6, 0.7, 0.8, 0.9]$ \textcolor{black}{are} selected as the enriched apex frames. Second, 5 frames in the position of ${\textit{Pos}}_{\textit{apex}} + ({\textit{Pos}}_{\textit{offset}} - {\textit{Pos}}_{\textit{apex}}) \times [0.1, 0.2, 0.3, 0.4, 0.5]$ \textcolor{black}{are} selected as the enriched apex frames. Finally, all  original onset/apex frames and enriched apex frames \textcolor{black}{are} rotated between the angles in $[-15^{\circ}, 15^{\circ}]$ with an increment of $5^{\circ}$. Performing these three strategies jointly, the original data \textcolor{black}{are} augmented by 70 times. Based on enriched data, we \textcolor{black}{calculate} the TV-L1 optical flow from the pair of original onset and enriched apex frames.

\textbf{Parameter setup:} Stochastic gradient descent with ADAM \textcolor{black}{is} used to learn the network parameters, where $\beta_1 = 0.9$, $\beta_2 = 0.999$, and $\epsilon = 10^{-8}$. The network \textcolor{black}{is} trained in 50 epochs by using  mini-batch size of 32 and a learning rate of 0.0005. The margin $\beta$ is set as  0.02. The loss weights $\lambda_1 = 1$ and $\lambda_2 = 0.2$ \textcolor{black}{are} used. All models \textcolor{black}{are} trained on a NVIDIA TITAN X GPU based on the Pytorch deep learning framework.

\subsection{Experiment on synthesized occluded micro-expression databases}
\label{sec:experimentD}
 For evaluating the robustness of our proposed method and the existing micro-expression features to micro-expression occlusion problem, we use THREE representative handcrafted descriptors and TWO deep learning networks for comparison. For three hand-engineered features, SVM is served as the classifier. \textcolor{black}{The results are represented in Tables \ref{occ_hde} and~\ref{occ_cde_result}.}

\textbf{(1) Comparison with handcrafted features}

We \textcolor{black}{compare} our RRRN with LBP-TOP~\cite{Zhao2007}, LBP-SIP~\cite{Wang2014b}, and 3DHOG~\cite{Polikovsky2009} on the synthesized occluded samples. Their parameters setups are the same as that of experiment on un-occluded micro-expression databases and SVM is served as the classifier. Tables~\ref{occ_hde} and~\ref{occ_cde_result} report the comparative results of three handcrafted features and our RRRN on HDE and CDE tasks, respectively. In Table~\ref{occ_hde}, WAR and UAR are used, and in Table~\ref{occ_cde_result}, F1 and WF1 used.  \textcolor{black}{As shown in these two tables, some handcrafted features have very limited categorization capabilities, \ie,  on the fold of \emph{CASME II$\rightarrow$SAMM} of HDE task, LBP-TOP and LBP-SIP almost \textcolor{black}{categorize} all the samples into the dominant class where the WAR and UAR are 0.294 is 0.200, respectively.} It is also can be seen that our RRRN achieves significant increases in the performance over these features in all the occlusion experiments (Mask, Glass, and random occlusion). For example, the RRRN \textcolor{black}{obtains} the average WAR / UAR of 0.538 / 0.480 and 0.642 / 0.509 in the fold of \emph{CASME II$\rightarrow$SAMM} and \emph{SAMM $\rightarrow$ CASME II}, which are much higher than the results of LBP-TOP (0.293 / 0.199 and  0.334 / 0.217) in the two folds of HDE task, respectively. Additionally, the considerable improvement is observed by comparing RRRN and LBP-TOP in the CDE task. The experimental results indicate RRRN can learn more robust representation than three handcrafted features for MER. \textcolor{black}{Moreover, the results} demonstrate the effectiveness of deep learning methods.

%
%

\textbf{(2) Comparison with deep learning methods}

We compared RRRN with ResNet18 and VGG16, where The ResNet18 is pre-trained on ImageNet~\cite{Deng2009} and VGG16 downloaded from website \footnote{https://www.robots.ox.ac.uk/~vgg/software/vgg\_face/}. The comparison in HDE and CDE tasks are shown in Tables~\ref{occ_hde} and~\ref{occ_cde_result}. The average performance of RRRN is significantly better than that of VGG16 and ResNet18. Under nearly all the occlusion conditions, RRRN has little performance degradation while VGG16 and ResNet18 suffer more performance degradation.
In detail, in the fold of \emph{CASME II$\rightarrow$SAMM} task, RRRN obtains the maximal WAR degradation of 17.7\%, while VGG16 and ResNet18 19.1\% and 17.8\% amongst the 12 occlusion experiments. Additionally, in the CDE task, the maximal WF1 degradation of RRRN is 21.5\%, while that of VGG16 and ResNet18 26.5\% and 23.3\%, respectively.

The proceeds of RRRN, especially when compared with ResNet18, are due to the significant roles of RI module and the RR module in RRRN. As RI module explores the weighted region feature by using the attention mechanism, RRRN has capability to capture the subtle muscle motions in facial regions and discover the micro-expression related regions and existed occluded facial regions. By the relational feature learning in the RR module, RRRN has a more robust representation with the complementary information in different facial regions.

\begin{table*}
\caption{Performance comparison on the HDE task with synthesized occluded samples (WAR/UAR). The best results are highlighted in bold.} \label{all_results}
\label{occ_hde}
\begin{center}
\subtable[Performance comparison on the fold of \emph{CASME II$\rightarrow$SAMM}.]{
\label{occ_samm_result}
\setlength{\tabcolsep}{1mm}{
\begin{tabular}{|c|c|c|c|c|c|c||c|}
\hline
Occ&LBP-TOP \cite{Zhao2007}&LBP-SIP \cite{Wang2014b}&3DHOG \cite{Polikovsky2009}&VGG16 \cite{Simonyan2014}&ResNet18 \cite{He2016a}&\textbf{RRRN (Ours)}&Avg.\\
\hline	
\hline	
\emph{Un-occ}  &0.294 / 0.200 & 0.294 / 0.200 & 0.353 / 0.227 & 0.591 / 0.542 & 0.603 / 0.530 & \textbf{0.647 / 0.559} & 0.464 / 0.376\\\hline
\emph{Mask}    &0.324 / 0.213 &	0.294 /	0.200 &	0.294 /	0.193 &	0.369 /	0.306 &	0.416 /	0.337 &	\textbf{0.471 /	0.361} &	0.361 /	0.268\\\hline
\emph{Glass}   &0.294 /	0.200 &	0.294 /	0.200 &	0.265 /	0.175 &	0.350 /	0.313 &	0.380 /	0.343 &	\textbf{0.409 /	0.396} &	0.332 /	0.271\\\hline
\emph{Occ-5\%} &0.294 /	0.200 &	0.294 /	0.200 &	0.221 /	0.202 &	0.541 /	0.474 &	0.558 /	0.483 &	\textbf{0.586 /	0.515} &	0.416 /	0.346\\\hline
\emph{Occ-10\%}&0.294 /	0.200 &	0.294 /	0.200 &	0.206 /	0.137 &	0.518 /	0.468 &	0.533 /	0.464 &	\textbf{0.580 /	0.512} &	0.404 /	0.330\\\hline
\emph{Occ-15\%}&0.294 /	0.200 &	0.294 /	0.200 &	0.235 /	0.160 &	0.474 /	0.433 &	0.513 /	0.443 &	\textbf{0.557 /	0.496} &	0.395 /	0.322\\\hline
\emph{Occ-20\%}&0.294 /	0.200 &	0.294 /	0.200 &	0.279 /	0.200 &	0.493 /	0.448 &	0.520 /	0.452 &	\textbf{0.559 /	0.485} &	0.407 /	0.331\\\hline
\emph{Occ-25\%}&0.294 /	0.200 &	0.294 /	0.200 &	0.279 /	0.200 &	0.480 /	0.441 &	0.496 /	0.430 &	\textbf{0.561 /	0.493} &	0.401 /	0.327\\\hline
\emph{Occ-30\%}&0.279 /	0.190 & 0.294 / 0.200 &	0.235 /	0.173 &	0.460 /	0.419 &	0.478 /	0.431 &	\textbf{0.561 /	0.493} &	0.385 /	0.318\\\hline
\emph{Occ-35\%}&0.294 /	0.200 &	0.294 /	0.200 &	0.206 /	0.145 &	0.475 /	0.450 &	0.501 /	0.450 &	\textbf{0.553 /	0.515} &	0.387 /	0.327\\\hline
\emph{Occ-40\%}&0.279 /	0.190 &	0.294 /	0.200 &	0.250 /	0.170 &	0.138 /	0.414 &	0.474 /	0.401 &	\textbf{0.504 /	0.455} &	0.323 /	0.305\\\hline
\emph{Occ-45\%}&0.279 /	0.190 &	0.294 /	0.200 &	0.206 /	0.155 &	0.431 /	0.440 &	0.458 /	0.419 &	\textbf{0.511 /	0.486} &	0.363 /	0.315\\\hline
\emph{Occ-50\%}&0.294 /	0.200 &	0.294 /	0.200 &	0.265 /	0.293 &	0.441 /	0.457 &	0.463 /	0.416 &	\textbf{0.500 /	0.469} &	0.376 /	0.339\\\hline\hline
Avg.           &0.293 /	0.199 &	0.294 /	0.200 &	0.253 /	0.187 &	0.443 /	0.431 &	0.492 /	0.431 &	\textbf{0.538} / \textbf{0.480} &	-	\\
\hline
\end{tabular}}
\label{occ_samm_result}
}
\subtable[Performance comparison on the fold of \emph{SAMM $\rightarrow$ CASME II}.]{
\label{occ_casme2_result}
\setlength{\tabcolsep}{1mm}{
\begin{tabular}{|c|c|c|c|c|c|c||c|}
\hline
Occ&LBP-TOP \cite{Zhao2007}&LBP-SIP \cite{Wang2014b}&3DHOG \cite{Polikovsky2009}&VGG16 \cite{Simonyan2014}&ResNet18 \cite{He2016a}&\textbf{RRRN (Ours)}&Avg.\\
\hline	
\hline		
\emph{Un-occ}  &0.465 /	0.255 &	0.157 /	0.208 &	0.374 /	0.263 &	0.629 /	0.533 &	0.659 /	0.540 &	\textbf{0.700 /	0.592} &	0.497 /	0.399\\\hline
\emph{Mask}    &0.405 /	0.170 &	0.297 /	0.237 &	0.178 /	0.181 &	0.440 /	0.362 &	0.487 /	0.343 &	\textbf{0.557 /	0.399} &	0.394 /	0.282\\\hline
\emph{Glass}   &0.314 /	0.224 &	0.141 /	0.202 &	0.314 /	0.192 &	0.410 /	0.370 &	0.445 /	0.396 &	\textbf{0.470 /	0.399} &	0.349 /	0.297\\\hline
\emph{Occ-5\%} &0.422 /	0.217 &	0.151 /	0.206 &	0.395 /	0.321 &	0.609 /	0.524 &	0.664 /	0.522 &	\textbf{0.677 /	0.418} &	0.486 /	0.368\\\hline
\emph{Occ-10\%}&0.422 /	0.251 &	0.184 /	0.212 &	0.378 /	0.298 &	0.596 /	0.520 &	0.639 /	0.502 &	\textbf{0.668 /	0.562} &	0.481 /	0.391\\\hline
\emph{Occ-15\%}&0.395 /	0.237 &	0.216 /	0.218 &	0.405 /	0.268 &	0.597 /	0.499 &	0.651 /	0.502 &	\textbf{0.672 /	0.548} &	0.489 /	0.379\\\hline
\emph{Occ-20\%}&0.335 /	0.215 &	0.184 /	0.206 &	0.351 /	0.214 &	0.601 /	0.515 &	0.648 /	0.506 &	\textbf{0.681 /	0.547} &	0.467 /	0.367\\\hline
\emph{Occ-25\%}&0.335 /	0.215 &	0.184 /	0.206 &	0.351 /	0.214 &	0.589 /	0.494 &	0.638 /	0.493 &	\textbf{0.669 /	0.562} &	0.461 /	0.364\\\hline
\emph{Occ-30\%}&0.287 /	0.221 &	0.238 /	0.232 &	0.308 /	0.178 &	0.581 /	0.484 &	0.636 /	0.477 &	\textbf{0.650 /	0.548} &	0.450 /	0.357\\\hline
\emph{Occ-35\%}&0.297 /	0.213 &	0.222 /	0.202 &	0.411 /	0.235 &	0.590 /	0.494 &	0.630 /	0.480 &	\textbf{0.659 /	0.517} &	0.468 /	0.357\\\hline
\emph{Occ-40\%}&0.227 /	0.204 &	0.184 /	0.194 &	0.465 /	0.285 &	0.578 /	0.475 &	0.628 /	0.474 &	\textbf{0.658 /	0.526} &	0.457 /	0.360\\\hline
\emph{Occ-45\%}&0.238 /	0.215 &	0.227 /	0.210 &	0.432 /	0.227 &	0.572 /	0.468 &	0.617 /	0.451 &	\textbf{0.648 /	0.515} &	0.456 /	0.348\\\hline
\emph{Occ-50\%}&0.205 /	0.190 &	0.260 /	0.211 &	0.378 /	0.220 &	0.562 /	0.447 &	0.616 /	0.446 &	\textbf{0.632 /	0.486} &	0.442 /	0.333\\\hline\hline
Avg.           &0.334 /	0.217 &	0.203 /	0.211 &	0.365 /	0.238 &	0.566 /	0.476 &	0.612 /	0.472 &	\textbf{0.642} / \textbf{0.509} &	 -	\\
\hline
\end{tabular}}
\label{occ_casme2_result}
}
\end{center}
\end{table*}

\begin{table*}
\caption{Performance comparison on on the CDE task with synthesized occluded samples (F1/WF1). The best results are highlighted in bold.} \label{all_results}
\label{occ_cde_result}
\begin{center}
\setlength{\tabcolsep}{1mm}{
\begin{tabular}{|c|c|c|c|c|c|c||c|}
\hline
Occ&LBP-TOP \cite{Zhao2007}&LBP-SIP \cite{Wang2014b}&3DHOG \cite{Polikovsky2009}&VGG16 \cite{Simonyan2014}&ResNet18 \cite{He2016a}&\textbf{RRRN (Ours)}&Avg.\\
\hline
\hline		
\emph{Un-occ}  &0.146 /	0.096 &	0.145 /	0.095 &	0.244 /	0.188 &	0.591 /	0.687 &	0.641 /	0.717 &	\textbf{0.663 /	0.735} &	0.405 /	0.420\\\hline
\emph{Mask}    &0.134 /	0.184 &	0.129 /	0.086 &	0.200 /	0.169 &	0.327 /	0.467 &	0.316 /	0.464 &	\textbf{0.377 /	0.497} &	0.247 /	0.311\\\hline
\emph{Glass}   &0.181 /	0.114 &	0.167 /	0.106 &	0.179 /	0.156 &	0.409 /	0.524 &	0.420 /	0.524 &	\textbf{0.447 /	0.554} &	0.301 /	0.330\\\hline
\emph{Occ-5\%} &0.168 /	0.108 &	0.135 /	0.090 &	0.177 /	0.230 &	0.581 /	0.672 &	0.623 /	0.697 &	\textbf{0.637 /	0.712} &	0.387 /	0.418\\\hline
\emph{Occ-10\%}&0.189 /	0.120 &	0.133 /	0.089 &	0.182 /	0.139 &	0.543 /	0.638 &	0.576 /	0.660 &	\textbf{0.591 /	0.668} &	0.369 /	0.3861\\\hline
\emph{Occ-15\%}&0.202 /	0.127 &	0.148 /	0.097 &	0.205 /	0.157 &	0.530 /	0.625 &	0.565 /	0.656 &	\textbf{0.593 /	0.675} &	0.374 /	0.390\\\hline
\emph{Occ-20\%}&0.200 /	0.125 &	0.151 /	0.098 &	0.192 /	0.157 &	0.526 /	0.623 &	0.557 /	0.653 &	\textbf{0.588 /	0.668} &	0.369 /	0.387\\\hline
\emph{Occ-25\%}&0.200 /	0.125 &	0.151 /	0.098 &	0.192 /	0.157 &	0.506 /	0.623 &	0.506 /	0.625 &	\textbf{0.542 /	0.641} &	0.350 /	0.378\\\hline
\emph{Occ-30\%}&0.161 /	0.100 &	0.167 /	0.107 &	0.189 /	0.170 &	0.474 /	0.587 &	0.516 /	0.615 &	\textbf{0.539 /	0.626} &	0.341 /	0.368\\\hline
\emph{Occ-35\%}&0.152 /	0.094 &	0.168 /	0.130 &	0.184 /	0.157 &	0.489 /	0.594 &	0.515 /	0.603 &	\textbf{0.534 /	0.628} &	0.340 /	0.368\\\hline
\emph{Occ-40\%}&0.101 /	0.060 &	0.169 /	0.147 &	0.194 /	0.170 &	0.457 /	0.583 &	0.510 /	0.608 &	\textbf{0.523 /	0.612} &	0.326 /	0.363\\\hline
\emph{Occ-45\%}&0.098 /	0.058 &	0.187 /	0.149 &	0.162 /	0.140 &	0.444 /	0.577 &	0.462 /	0.572 &	\textbf{0.496 /	0.594} &	0.308 /	0.348\\\hline
\emph{Occ-50\%}&0.089 /	0.052 &	0.197 /	0.175 &	0.172 /	0.107 &	0.415 /	0.541 &	0.443 /	0.556 &	\textbf{0.463 /	0.577} &	0.297 /	0.335\\\hline\hline
Avg.           &0.155 /	0.105 &	0.157 /	0.113 &	0.190 /	0.161 &	0.484 /	0.595 &	0.512 /	0.612 &	\textbf{0.538} / \textbf{0.630} &	 -	\\
\hline
\end{tabular}}
\end{center}
\end{table*}

\section{CONCLUSION}
In this paper, we go deep into the more challenging task in micro-expression recognition (MER), \ie, occlusion micro-expression recognition. We propose a novel approach for occlusion MER, named Region-inspired Relation Reasoning Network (RRRN), which involves three key components: the Backbone network, the Region-Inspired (RI) module, and the  Relation Reasoning (RR) module. The proposed method can automatically capture the subtle movement of micro-expressions in facial regions and learn the important factors for different facial regions according to the contribution of the regions for the final classification or according to the presence of occlusion.
Through reasoning the relationship between the input of the weighted region features, the model achieves robust representation with the aggregation of complementary region features. Experimental results on RRRN under non-occlusion and occlusion scenarios on two widely used CASME II and SAMM databases and our synthesized versions demonstrate that the proposed approach achieves high-recognition performance on different occlusions. \textcolor{black}{In} the future, we will study an end-to-end
approach for occlusion MER, as our model is based on pre-computed optical flow. \textcolor{black}{Moreover, we will} find more effective attention methods to explore subtle micro-expression movements, and introduce AUs relationship in micro-expressions for occlusion MER.

\ifCLASSOPTIONcaptionsoff
  \newpage
\fi


\end{document}